\pdfoutput=1

\documentclass[11pt]{article}

\usepackage[]{acl}
\usepackage{times}
\usepackage{latexsym}
\usepackage{amsmath}
\usepackage{booktabs}
\usepackage{amssymb}
\usepackage[T1]{fontenc}

\usepackage[utf8]{inputenc}

\usepackage{microtype}
\usepackage{multirow}
\title{OCHADAI at SemEval-2022 Task 2: Adversarial Training for Multilingual Idiomaticity Detection}

\author{Lis Kanashiro Pereira\textsuperscript{1}, Ichiro Kobayashi\textsuperscript{2} \\
  Ochanomizu University \\
  \texttt{kanashiro.pereira@ocha.ac.jp\textsuperscript{1}, koba@is.ocha.ac.jp\textsuperscript{2}} \\}

\begin{document}
\maketitle

\begin{abstract}

We propose a multilingual adversarial training model for determining whether a sentence contains an idiomatic expression. 
Given that a key challenge with this task is the limited size of annotated data, our model relies on pre-trained contextual representations from different multilingual state-of-the-art transformer-based language models (i.e., multilingual BERT and XLM-RoBERTa), and on adversarial training, a  training method for further enhancing model generalization and robustness.
Without relying on any human-crafted features, knowledge bases, or additional datasets other than the target datasets, our model achieved competitive results and ranked 6\textsuperscript{th} place in SubTask A (zero-shot) setting and 15\textsuperscript{th} place in SubTask A (one-shot) setting.
\end{abstract}

\section{Introduction}

Large-scale pre-trained language models such as BERT \cite{devlin2019bert} have achieved great success in a wide range of natural language processing (NLP) tasks. However, more recent studies show that even such contextual models have a limited ability to capture idiomaticity \cite{garcia2021probing}. Idiomatic expressions denote a group of words that behave as single
words to some extent. Their linguistic behavior cannot be inferred from the characteristics of their components, and still pose a challenge to natural language processing (NLP) systems.

This paper describes the system developed by the OCHADAI team for SemEval-2022 Task 2 - Multilingual Idiomaticity Detection and Sentence Embedding \cite{tayyarmadabushi-etal-2022-semeval}. Given that a key challenge in this task is the limited size of annotated data, we follow best practices from recent work on enhancing model generalization and robustness and propose a model ensemble that leverages multilingual pre-trained representations and adversarial training. Our model ranked 6th on SubTask A (zero-shot), and 15th on SubTask A (one-shot).
\section{Task Description}

SemEval-2021 Task 2 SubTask A consists of a binary classification task that requires classifying sentences with a target multiword expression (MWE) into either "Idiomatic" or "Literal" across  English,  Portuguese  and Galician \cite{tayyar-madabushi-etal-2021-astitchinlanguagemodels-dataset}. Further, it is subdivided into two settings to better test models' ability to generalize: zero-shot and one-shot. In the zero-shot" setting, multiword expressions (potentially idiomatic phrases), in the training set are completely disjoint from those in the test and development sets. In the "one-shot" setting, one positive and one negative training examples are included for each MWE in the test and development sets. Note that the actual examples in the training data are different from those in the test and development sets in both settings. Only the datasets provided by the organizers are allowed to train the models. Participants can use only the data provided for the zero-shot setting to train the zero-shot model. However, participants were allowed to use data provided for both settings to train models in the one-shot setting. The statistics of the corpus are presented in Table~\ref{data1-table}. Our team submitted results for both settings, and the next section outlines the overview of our model.

\begin{table}
\small
\centering
\begin{tabular}{l|l|c|c|c|c} 
\toprule
 \bf{Setting} & \bf{Language}& \bf{Train} & \bf{Dev} & \bf{Eval} & \bf{Test}\\ \hline
 \multirow{3}{1.3cm}{zero-shot} &English &  3,327         &   -   & -&  -     \\
                                &Portuguese &    1,164    &   -    & -&  -   \\ 
                                &Galician  &    0        &  -   &  -  & - \\ \hline
\multirow{4}{1.3cm}{one-shot}&English      &    87       &  466   &  483 & 916    \\
                             &Portuguese  &    53       &    273  & 279  &  713 \\
                             &Galician    &    0        &  0    &  0    & 713\\ 

\bottomrule
\end{tabular}
\caption{\label{data1-table} Summary of the SemEval 2022 Task 2 Subtask A dataset. Note that the dev, eval and test sets are used in both settings.}
\end{table}

\begin{table*}[!htb]
\begin{center}
\small
\begin{tabular}{@{\hskip2pt}c@{\hskip2pt}|@{\hskip2pt}c@{\hskip2pt}|p{10cm}|@{\hskip2pt}c@{\hskip2pt}|@{\hskip2pt}c@{\hskip2pt}}
\toprule

\bf{Setting}& \bf{Language}& \bf{Sentence} & \bf{Target MWE} & \bf{Label} \\ \hline

zero-shot& English & This song is about unconditionally supporting someone you love. This is a \textbf{love song}. Let’s be there for each other. & love song & 1   \\ 
one-shot&  Portuguese & Estamos honrando o teto e construindo as paredes”, afirmou. Em outro momento, voltando ao tema do impasse do Orçamento, ele reforçou que a busca é por uma solução, que a briga tem “mérito” e que “sempre, em grande rebanho, tem uma \textbf{ovelha negra}”."Mas não é o Congresso nem o grosso do ministério”, completou. & ovelha negra& 0  \\ 
one-shot& Galician & Non podemos abandonalos á súa sorte, porque sen mozos e mozas no campo e sen a súa actividade agraria suporía, entre outras cousas, a deslocalización da produción e a dependencia alimentaria Máis alá das políticas comunitarias, ¿cres que poden desenvolverse alternativas para a \textbf{xente nova} a partir de medidas municipais, autonómicas ou estatais? Estase a facer? & xente nova & 0   \\
\bottomrule
 
\end{tabular}
\caption{\label{tab:error_samples} Example sentences and labels for Subtask A.
Note that "Idiomatic" is assigned the label 0 in the dataset and "non-idiomatic" (including proper nouns) are assigned the label 1.}
\end{center}
\end{table*}

\section{System Overview}
\label{sec:model}
We focus on exploring different training techniques using BERT and RoBERTa, given their superior performance on a wide range of NLP tasks. Each text encoder and training method used in our model are detailed below.

\subsection{Text Encoders}
\noindent\textbf{M-BERT} \cite{devlin2019bert}: We use the M-BERT\textsubscript{BASE} model released by the authors. It is pre-trained on the top 104 languages with the largest Wikipedia using a masked language modeling (MLM) objective. This model is case sensitive: it makes a difference between english and English.

\noindent\textbf{XLM-R} \cite{conneau2019unsupervised}: XLM-RoBERTa (XLM-R) is a multilingual version of RoBERTa. It is pre-trained on 2.5TB of filtered CommonCrawl data containing 100 languages. XLM-R
has been shown to perform particularly well on low-resource languages, such as Swahili and Urdu. We use the XLM-R\textsubscript{LARGE} model released by the authors.

\subsection{Training Procedures}
\noindent\textbf{Standard fine-tuning}: 
This is the standard fine-tuning procedure where we fine-tune BERT and RoBERTa on each training setting-specific data.

\noindent\textbf{Adversarial training (ADV)}: Adversarial training has proven effective in improving model generalization and robustness in computer vision \cite{madry2017pgd,goodfellow2014explaining} and more recently in NLP \cite{zhu2019freelb,jiang2019smart,cheng2019robust,liu2020alum,pereira2020alic}. It works by augmenting the input with a small perturbation that maximizes the adversarial loss:
\begin{equation}
\min_{\theta} \mathbb{E}_{(x, y)\sim D}[\max_{ \delta} l(f(x + \delta; \theta), y)]
\label{eq:adv}
\end{equation}
where the inner maximization can be solved by projected gradient descent \cite{madry2017pgd}. Recently, adversarial training has been successfully applied to NLP as well \cite{zhu2019freelb,jiang2019smart,pereira2020alic}. In our experiments, we use SMART \cite{jiang2019smart}, which instead regularizes the standard training objective using {\em virtual adversarial training} \cite{miyato2018virtual}:
\begin{equation}
\begin{aligned}
  \min_{\theta} \mathbb{E}_{(x, y)\sim D}[l(f(x; \theta), y) + \\ \alpha  \max_{
\delta} l(f(x+\delta; \theta), f(x; \theta))]
\end{aligned}
\vspace{-1mm}
\label{eq:alum}
\end{equation}
Effectively, the adversarial term encourages smoothness in the input neighborhood, and $\alpha$ is a hyperparameter that controls the trade-off between standard errors and adversarial errors.
\section{Experiments}
\subsection{Implementation Details}
Our model implementation is based on
the MT-DNN framework \cite{liu2019multi,liu2020mtmtdnn}. We use BERT \cite{devlin2019bert} and XLM-R \cite{conneau2019unsupervised} as the text encoders.
We used ADAM \cite{kingma2014adam} as our optimizer with a learning rate in the range $\in \{8 \times 10^{-6}, 9 \times 10^{-6}, 1 \times 10^{-5}\}$ and a batch size $\in \{8, 16, 32\}$. The maximum number of epochs was set to 10. A linear learning rate decay schedule with warm-up over 0.1 was used, unless stated otherwise. To avoid gradient exploding, we clipped the gradient norm within 1. All the texts were tokenized using wordpieces and were chopped to spans no longer than 512 tokens. During adversarial training, we follow \cite{jiang2019smart} and set the perturbation size to $1 \times 10^{-5}$, the step size to $ 1 \times 10^{-3}$, and  to $ 1 \times 10^{-5}$ the variance
for initializing the perturbation. The number of projected gradient steps and the $\alpha$ parameter (Equation 2) were both set to $1$.

We follow \cite{devlin2019bert} and \cite{liu2019roberta}, and set the first token as the [CLS] token  and the <s> token, respectively, when encoding the input on BERT and RoBERTa, respectively. We separate the input sentence and the target expression with the special token [SEP] and </s> for BERT and RoBERTa, respectively. e.g. [CLS] Ben Salmon is a committed \textit{night owl} with an undying devotion to discovering new music.,"He lives in the great state of Oregon, where he hosts a killerradio show and obsesses about Kentucky basketball from afar. [SEP] \textit{night owl} [SEP].  

For both settings (zero-shot and one-shot), we used the dev dataset released by organizers to fine-tune the model's hyperparameters. 

\subsection{Main Results}

Submitted systems were evaluated in terms of F1-score. The systems were ranked from highest F1-score score to lowest. We built several models that use different text encoders and different training methods, as described in Section \ref{sec:model}. See \autoref{tab:main} for the results. 

First, we observe that models that use adversarial training obtained better performance overall, without using any additional knowledge source, and without using any additional dataset other than the target task datasets. These results suggest that adversarial training leads to a more robust model and helps generalize better on unseen data. For the zero-shot setting, the model that uses XLM-R as the text encoder and adversarial training performed better than M-BERT on the development set. Thus, we decided to submit this model's results on the test set. It obtained a test set F1-score of 0.7457, and ranked 6\textsuperscript{th} among all participating systems. On the other hand, on the one-shot setting, M-BERT performed better than XLM-R on the development set. Again, M-BERT with adversarial training performed better than vanilla fine-tuning. This model obtained an F1-score of 0.6573 on the test set, and ranked 15\textsuperscript{th} among all participating systems.

\begin{table}[!htb]
    \centering
    \begin{tabular}{l|c|c}
    \hline
\toprule    
         \multirow{2}{*}{\bf Method} 
          & \bf{zero-shot} & \bf{one-shot} \\
          & F1 & F1 \\ \hline
          \multicolumn{2}{r}{Dev} \\ \hline 
          Standard\textsubscript{XLM-R\textsubscript{LARGE}} &0.7076	&  0.7769\\ 
          SMART\textsubscript{XLM-R\textsubscript{LARGE}} & 0.7378	&  0.7958 \\ \hline  
          Standard\textsubscript{M-BERT\textsubscript{BASE}} &0.5540	& 0.8462  \\ 
          SMART\textsubscript{M-BERT\textsubscript{BASE}} &0.6200	& 0.8568 \\ \hline 
          \multicolumn{2}{r}{Test} \\ \hline 
          SMART\textsubscript{XLM-R\textsubscript{LARGE}} &0.7457 &  - \\ 
          SMART\textsubscript{M-BERT\textsubscript{BASE}} & -	&  0.6573 \\  
          
\bottomrule
    \end{tabular}
    \caption{Comparison of standard and adversarial training in zero-shot evaluation on various natural language inference datasets, where the standard BERT\textsubscript{BASE} model is fine-tuned on the MNLI training data.}
    \label{tab:main}
\end{table}

\section{Conclusion}
\vspace{-2mm}
We proposed a simple and efficient model for multilingual idiomaticity detection. Our experiments demonstrated that it achieves competitive results on both zero-shot and one-shot settings, without relying on any additional resource other than the target task dataset. Although in this paper we focused on the multilingual idiomaticity detection task, our model can be generalized to solve other downstream tasks as well, and we will explore this direction as future work.

\section*{Acknowledgements}
We thank the reviewers for their helpful feedback. This work has been supported by the project KAKENHI ID: 21K17802.

\bibliography{anthology,custom}
\bibliographystyle{acl_natbib}




\end{document}